\pdfoutput=1
\documentclass[11pt]{article}

\usepackage{EMNLP2022}

\usepackage{times}
\usepackage{latexsym}

\usepackage[T1]{fontenc}
\usepackage[utf8]{inputenc}

\usepackage{microtype}

\usepackage{comment}
\usepackage{float}
\usepackage{xcolor}
\restylefloat{table}
\usepackage{amsmath}
\usepackage[utf8]{inputenc}
\usepackage{graphicx}

\usepackage{booktabs}
\usepackage{multirow}
\usepackage{stmaryrd}
\usepackage{savesym}
\usepackage{dingbat}
\newcommand{\best}[1]{{\color{red}#1}}

\makeatletter
\def\thickhline{%
  \noalign{\ifnum0=`}\fi\hrule \@height \thickarrayrulewidth \futurelet
   \reserved@a\@xthickhline}
\def\@xthickhline{\ifx\reserved@a\thickhline
               \vskip\doublerulesep
               \vskip-\thickarrayrulewidth
             \fi
      \ifnum0=`{\fi}}
\makeatother

\newlength{\thickarrayrulewidth}
\title{How Far are We from Robust Long Abstractive Summarization?}

\author{Huan Yee Koh$^{1}$\thanks{\hspace{2mm}Equal contribution.} , Jiaxin Ju$^{4}$\footnotemark[1] , He Zhang$^{5}$, Ming Liu$^{2}$\thanks{\hspace{2mm}Corresponding author.} , Shirui Pan$^{3}$\footnotemark[2] \\
$^1$Faculty of Information Technology, Monash University, Australia\\
 $^2$School of Information Technology, Deakin University, Australia\\
 $^3$School of Information and Communication Technology, Griffith University, Australia \\
 $^4$Independent Researcher\\
 $^5$Zhongtukexin Co. Ltd., Beijing, China \\
 \texttt{huan.koh@monash.edu, jiaxin.ju.14@gmail.com, zhanghe@kxsz.net}\\
  \texttt{m.liu@deakin.edu.au, }
  \texttt{s.pan@griffith.edu.au}\\
}

\begin{document}
\maketitle
\begin{abstract}
Abstractive summarization has made tremendous progress in recent years. In this work, we perform fine-grained human annotations to evaluate long document abstractive summarization systems (i.e., models and metrics) with the aim of implementing them to generate reliable summaries. For long document abstractive models, we show that the constant strive for state-of-the-art ROUGE results can lead us to generate more relevant summaries but not factual ones. For long document evaluation metrics, human evaluation results show that ROUGE remains the best at evaluating the relevancy of a summary. It also reveals important limitations of factuality metrics in detecting different types of factual errors and the reasons behind the effectiveness of BARTScore. We then suggest promising directions in the endeavor of developing factual consistency metrics. Finally, we release our annotated long document dataset with the hope that it can contribute to the development of metrics across a broader range of summarization settings. 
\end{abstract}
\section{Introduction}
Pre-trained Transformers \cite{devlin2018bert,raffel2019exploring} have brought tremendous progress in summarizing text in an abstract manner \cite{rothe2021thorough}. Unlike extractive summarization \cite{xiao2019extractive,cui2021sliding,ju2021leveraging,shi2022starsum}, abstractive summarization presents a blue-sky potential of generating summaries that are fluent and relevant to the source by intelligently paraphrasing salient contents rather than merely copying from source texts \cite{beltagy2020longformer,ju2020monash,zaheer2020big,huang2021efficient}. Nevertheless, even under a short document setting, Transformer-based abstractive models often generate summaries that are repetitive \cite{see2019massively,holtzman2019curious}, ungrammatical, and factually inconsistent with the source \cite{durmus2020feqa,kryscinski-etal-2020-evaluating,maynez2020faithfulness}. Furthermore, current pre-trained Transformers have an input length limit that restricts them to be directly adapted to long document summarization \cite{lewis2020bart,zhang2020pegasus} as it would lead to a significant loss of salient information in the remaining text. These naturally bring us to a question: \textit{How far are we from building a robust abstractive summarization system for long documents?} 

A robust abstractive summarization system should at least have (i) models that can generate high-quality summaries, and (ii) evaluation metrics that can critically assess the relevance and factuality of a summary\footnote{In machine learning parlance, robustness refers to the ability of a model to adapt to unseen distribution. Here, robustness refers to the effectiveness of a model to adapt from short to long documents to generate relevant and factual summaries.}. However, research on analysis and critiques of models \cite{wilber-etal-2021-point,ladhak-etal-2022-faithful} and metrics \cite{gabriel-etal-2021-go,pagnoni-etal-2021-understanding} mainly focus on the short-document \cite{kryscinski2019neural,fabbri2021summeval} or long dialogue \cite{zhang2021exploratory}. Consequently, our work aims to fill the gap by systematically analyzing abstractive models and evaluation metrics under the long document setting. 

To analyze the quality of current state-of-the-art long document abstractive models, we lack a set of model-generated summaries with sufficient diversity under long document settings. To this end, we implement BART \cite{lewis2020bart} and PEGASUS \cite{zhang2020pegasus} models under arXiv \cite{cohan2018discourse} and GovReport \cite{huang2021efficient} as they have been found to be the most effective pre-trained Transformer in a large-scale evaluation of summarization models \cite{fabbri2021summeval}. However, their 1,024 token input limit would lead to a significant loss in the information required to generate a high-quality summary. 

Hence, by closely following prior works in extending the pre-trained models using sparse attention \cite{beltagy2020longformer,zaheer2020big,huang2021efficient} and reduce-then-summarize mechanism \cite{pilault2020extractive,zhang2021summ}, we implement different variants of Longformer-based BART and PEGASUS to obtain a diverse set of summaries. We then perform fine-grained human analysis on the model outputs by three human annotators to qualitatively assess whether long document abstractive models can generate relevant and factually consistent summaries.  

Effective evaluation metrics are also paramount as they can critically assess the model performance before releasing it to target users. We adapt recently proposed metrics \cite{durmus2020feqa,kryscinski-etal-2020-evaluating,nan2021improving,yuan2021bartscore,laban2022summac} to long document settings and thoroughly analyze their strength and weaknesses to measure the relevance and factual consistency on our annotated dataset. To our best knowledge, we are the first to assess abstractive models and evaluation metrics under the long document setting. 

Our contributions are as follows: (1) We analyze pre-trained Transformer summarizers to encourage a rethinking of architectural designs under long document settings. (2) We release human-annotated long document abstractive model outputs to further research in human-correlated evaluation metrics across a broader setting. (3) We investigate summarization metrics using our annotated long document datasets to expose the limitation of metrics and provide promising directions for the future development of evaluation metrics. 
\section{Related Work}
\subsection{Long Abstractive Models}
To implement pre-trained Transformers \cite{devlin2018bert,raffel2019exploring} for long document summarization tasks, they have to be adapted with \textit{long document mechanisms} to improve models' efficiency and extend their input limit \cite{koh2022longsurvey}. In this work, we focus on analyzing abstractive models after incorporating the two following long document mechanisms: 

\paragraph{Sparse Attention} It aims to reduce the quadratic complexity of Transformers into sub-quadratic complexity \cite{child2019generating,kitaev2019reformer,choromanski2020rethinking} while exploiting the benefits of pre-training \cite{beltagy2020longformer,zaheer2020big,huang2021efficient,guo2021longt5,pietruszka-etal-2022-sparsifying}. The gain in efficiencies allows Transformer to be fine-tuned on downstream summarization tasks with a substantially longer input text. Despite a plethora of proposals on sparse attention, \citet{xiong2022simple} recently showed that simple local attention remains competitive. 

\paragraph{Reduce-then-Summarize} This approach aims to reduce the source text into a shorter subset, allowing it to fit within the input token limit of a Transformer. The source text can be reduced into a more condensed text through extraction of salient sentences \cite{pilault2020extractive,zhao2020seal,bajaj-etal-2021-long} or generation of shorter texts from segments of the source \cite{gidiotis2020divide,zhang2021summ}. These models often train Transformer-based summarizers using reduced source texts which greedily maximize ROUGE scores and utilize separate retrievers during the testing stage to avoid "cheating" \cite{pilault2020extractive,manakul2021longspan,mao-etal-2022-dyle}. Importantly, the retriever will also be trained to maximize ROUGE to avoid a significant disconnect between the training and testing stage.

\subsection{Evaluation Metrics} 
Given the limitations of the ROUGE metric \cite{chaganty-etal-2018-price,kryscinski2019neural}, new metrics are proposed to better measure two fundamental qualities of summary: relevance and factual consistency. Relevance metrics such as ROUGE variants \cite{ng2015better,ganesan2018rouge,shafieibavani2018graph} and BERTScore \cite{zhang2019bertscore} measure whether a summary contains the main ideas of the source. A factual consistency metric assesses whether a summary is factually consistent with the source  \cite{goyal-durrett-2020-evaluating,wang2020asking}. Due to the high rate of factual errors in the summaries generated by short-document models \cite{cao2018faithful,maynez2020faithfulness},  there have been substantial efforts in developing effective metrics which can measure the factuality of a summary \cite{honovich2021q2,xie2021factual,ribeiro2022factgraph}. 

\section{Generation of Model Summary}\label{exp_design}
To investigate the robustness of long document abstractive systems, we need a set of model-generated summaries that can roughly represent the state of current research progress. In this section, we describe our methodology to obtain such samples.  

\subsection{Model Variants} 
\paragraph{Pretraining Task} We implement BART \cite{lewis2020bart} and PEGASUS \cite{zhang2020pegasus}. Both models have a 1,024 input token limit with extra text tokens to be truncated. We extend the input limit of BART and PEGASUS using the sparse attention and reduce-then-summarize mechanism. 

\paragraph{Sparse Attention} We extend the input limit of the pre-trained Transformer using Longformer's adaptation to have a maximum input of 1K, 4K, and 8K tokens \cite{beltagy2020longformer}. \citet{xiong2022simple} recently showed that local-window attentions (i.e., only attending to neighborhood tokens) are sufficient and competitive against other variants. The Longformer sparse attention adaptation thus gives us a reasonable baseline representation for current long document abstractive summarizers.

\paragraph{Reduce-then-Summarize} To explore the effectiveness of the reduce-then-summarize approach, we implement an oracle retriever by greedily extracting salient sentences that maximize ROUGE-2 up to the input limit of Transformer during the training and inference stage. Although using reference summaries to extract the salient sentences at the testing stage is considered cheating, contemporary approaches are trained to retrieve oracle summaries and are thus trained to become an oracle retriever \cite{manakul2021longspan,mao-etal-2022-dyle}. Using an oracle retriever allows us to analyze whether the reduce-then-summarize approach will generate desirable summaries given that the retriever is \textit{perfectly} trained with its upper bound performance. This allows us to analyze whether the summary generated from a ROUGE-maximizing model with a reduce-then-summarize mechanism will be desirable for target users. We implement models with 1K, 4K, and 8K tokens of the reduced subset.

\begin{table*}%[!htb]
    \centering
    \scalebox{0.73}{
    \begin{tabular}{lr|ccc|c|ccc|ccc}
        \toprule 
        \parbox[t]{2mm}{\multirow{11}{*}{\rotatebox[origin=c]{90}{Sparse Attention}}} & {\multirow{2}{*}{\textbf{Summarization Model}}} &  {\multirow{2}{*}{\textbf{Longformer}}} & {\multirow{2}{*}{\textbf{InputType}}} & {\multirow{2}{*}{\textbf{InputLen}}} & {\multirow{2}{*}{\textbf{InfoLoss}}} & \multicolumn{3}{c|}{\textbf{arXiv}} & \multicolumn{3}{c}{\textbf{GovReport}}
        \\ & & & & & & \textbf{R-1} & \textbf{R-2}  & \textbf{R-L} & \textbf{R-1} & \textbf{R-2}  & \textbf{R-L} \\
        \midrule
        % \midrule
        & BART (LEAD 1K) & \textbf{x} & \textbf{LEAD} & \textbf{1K} & 80\% & 43.84 & 16.55  & 39.86  & 56.55  & 26.70 & 54.46 \\
        & BART (LEAD 4K) & \textbf{\checkmark} & \textbf{LEAD} & \textbf{4K}  & 40\% & 45.72 & 18.48  & 41.82  & 57.45  & 28.14 & 55.40 \\
        & BART (LEAD 8K) & \textbf{\checkmark} & \textbf{LEAD} & \textbf{8K} & 20\% & 46.60 & 19.05  & 42.21  & 58.35  & 28.78 & \textbf{56.35} \\
        \cmidrule{2-12}
        & PEGASUS (LEAD 1K) & \textbf{x} & \textbf{LEAD} & \textbf{1K}  & 80\% & 44.17 & 17.16 & 40.18  & 57.19  & 27.87 & 55.17 \\
        & PEGASUS (LEAD 4K) & \textbf{\checkmark} & \textbf{LEAD} & \textbf{4K} & 40\% & 46.02 & 18.33 & 42.28  & 58.35  & 28.78 & 56.35 \\
        & PEGASUS (LEAD 8K) & \textbf{\checkmark} & \textbf{LEAD} & \textbf{8K} & 20\% & \textbf{46.87} & \textbf{19.73} & \textbf{42.36} & \textbf{58.59} & \textbf{29.02} & 56.29 \\
        \toprule
        \parbox[t]{2mm}{\multirow{6}{*}{\rotatebox[origin=c]{90}{Reduce-then-Summ}}}
        & BART (ORACLE 1K) & \textbf{x} & \textbf{ORACLE}& \textbf{1K} & - & 50.43 & \best{\textbf{\underline{24.16}}} & 44.93 & 63.07 & 36.64 & 60.09 \\
        & BART (ORACLE 4K) & \textbf{\checkmark} & \textbf{ORACLE} & \textbf{4K}  & - & 49.75 & 23.05 & 44.41 & 60.21 & 31.34 & 57.13 \\
        & BART (ORACLE 8K) & \textbf{\checkmark} & \textbf{ORACLE} & \textbf{8K} & - & 49.13 & 21.52 & 44.72 & 59.06 & 29.66 & 56.37 \\
        \cmidrule{2-12}
        & PEGASUS (ORACLE 1K) & \textbf{x} & \textbf{ORACLE} & \textbf{1K}  & - & \textbf{50.50} & 23.59 & \textbf{45.03} & \best{\textbf{\underline{63.47}}} & \best{\textbf{\underline{37.27}}} & \best{\textbf{\underline{60.52}}} \\
        & PEGASUS (ORACLE 4K) & \textbf{\checkmark} & \textbf{ORACLE} & \textbf{4K} & - & 46.21 & 20.32 & 42.23 & 60.86 & 33.68 & 57.88 \\
        & PEGASUS (ORACLE 8K) & \textbf{\checkmark} & \textbf{ORACLE} & \textbf{8K} & - & 49.06 & 20.60 & 43.55 & 58.77 & 31.53 & 56.51 \\
        \toprule
        \parbox[t]{4mm}{\multirow{2}{*}{\rotatebox[origin=c]{90}{SOTA }}} 
        & TDT \cite{pang2022long} & \textbf{\#N/A} & \textbf{LEAD} & \textbf{16K} & - & \best{\textbf{\underline{50.95}}} & 21.93 & \best{\textbf{\underline{45.61}}} & - & - & - \\
        \cmidrule{2-12}
        & DYLE \cite{mao-etal-2022-dyle} & \textbf{\#N/A} & \textbf{DYNAMIC} & \textbf{\#N/A} & - & 46.41 & 17.95 & 41.54 & 61.01 & 28.83 & 57.82 \\
        \bottomrule
    \end{tabular}
    }
    \caption{ROUGE score validation of implemented pre-trained BART and PEGASUS. SOTA stands for current state-of-the-art on arXiv, TDT \cite{pang2022long}, and GovReport, DYLE \cite{mao-etal-2022-dyle}. \best{\textbf{\underline{Red}}} represents best dataset result and \textbf{Bold} represents best result under the sparse attention or reduce-then-summarize setting.}
    \label{tab. rouge_validation}
\end{table*}

\subsection{Long Document Dataset}\label{sec:info_loss}
We implement the model configurations above on the ArXiv \cite{cohan2018discourse} and GovReport \cite{huang2021efficient} because they cover a wide range of topics in the scientific and general domains respectively. Both have an average source length of greater than 6,000 tokens, sufficiently long to challenge pre-trained Transformers. Besides, arXiv requires models to paraphrase more as compared to GovReport. Both datasets are chosen after analyzing the characteristics of datasets across 10 benchmark datasets with details in Appendix \ref{dataset_comparison}. 

\citet{kryscinski2019neural} has shown that 60\% of most important sentences lie within the leading one-third of the CNN-DM articles \cite{nallapati2016abstractive}. However, the linguistic styling and structure of a short document would often differ significantly from a long document. To investigate how much information a model would lose when processing only the leading text, we plot the distribution of salient content of arXiv and GovReport. This is done by performing human annotation on 10\% randomly sampled document-summary pairs from arXiv (700) and GovReport (100) test set. For each sentence in the reference summaries, we trace back the leading source paragraph position in the original document that contains the idea required to generate a sentence. 

\begin{figure} [!h]
    \centering
       \includegraphics[width=0.48\textwidth]{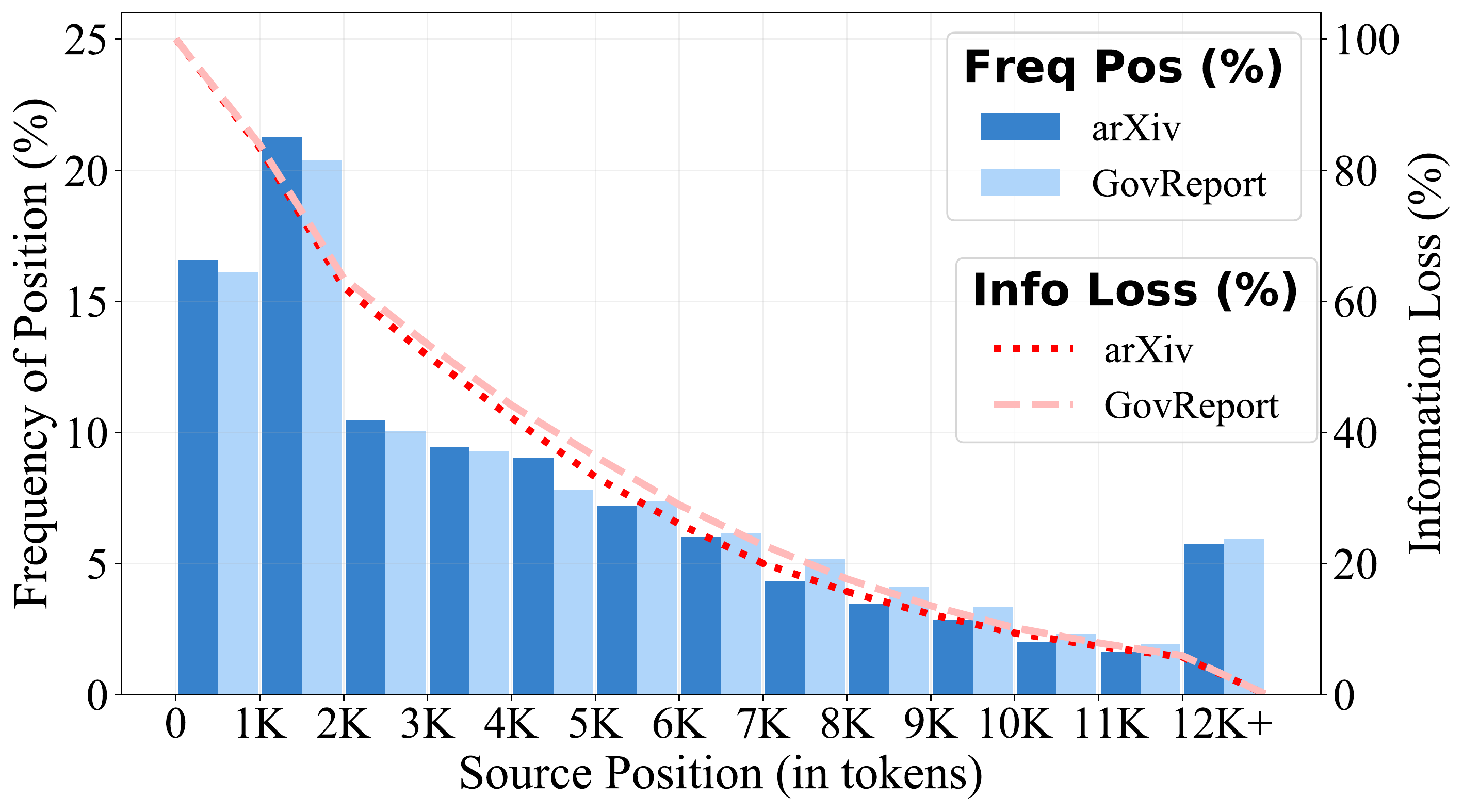}\\
       \vspace{-0.3cm}
       \caption{Distribution of salient content against the document length according to human annotators (left) ; Information loss of Transformer-based abstractive summarizers based on different input limits (right).}
       \label{fig. info_loss}
\end{figure}

Distribution plot in Figure \ref{fig. info_loss} shows the source position frequency in terms of the total percentage of the occurrence. The line plots illustrate the total information loss given an input limit. This reflects the information loss of a model when it only takes the leading source tokens. The line plot suggests that an input limit of 1K, 4K, and 8K tokens would equate to roughly 80\%, 40\%, and 20\% average information loss respectively on both datasets. 

Importantly, we see more salient information to be distributed from 1K to 2K tokens than 0 to 1K tokens, suggesting that the strategy of vanilla BART and PEGASUS to process the leading 1K input limit is sub-optimal. We hope that the result here would also provide directions for future architectural designs to identify salient contents. 

\subsection{Training Details}
Given two pre-training tasks with three input limit settings for Longformer-based Sparse Attention and Reduce-then-Summarize settings, this gives us 12 model configurations per dataset. For 1K token configurations, we use BART-large and PEGASUS-large. For 4K and 8K token configurations, we follow Longformer's implementation in extending the position embedding to 4K and 8K tokens by repeatedly copying position embeddings of BART and PEGASUS. To ensure comparability, all 24 models have a fixed output length of 512 tokens and are fine-tuned independently on RTX 3090 GPU with 24 GiB of GPU memory. We follow original authors in train/validation/test split of ArXiv \cite{cohan2018discourse} and GovReport \cite{huang2021efficient}. Implementation details in Appendix \ref{subsec:model_implementation}.

\subsection{ROUGE Validation}
Table \ref{tab. rouge_validation} shows that sparse attention models achieve competitive but lower ROUGE than state-of-the-art models, arXiv-TDT \cite{pang2022long} and GovReport-DYLE \cite{mao-etal-2022-dyle}. Extending the vanilla BART and PEGASUS using Longformer also provides a performance boost as the information loss is reduced exponentially when the input limit increased from 1K to 4K and 8K. The reduce-then-summarize models achieve ROUGE that either match or exceed arXiv's and GovReport's state-of-the-art. As increasing the input length would place more burden on reduce-then-summarize models to identify tokens that maximize ROUGE over long sequences, we see a slight decrease in ROUGE as the length is increased.  

The above results indicate that the implemented Longformer-based sparse attention models can reasonably reflect the current long abstractive summarization baselines, while the reduce-then-summarize models can roughly represent the summary outputs of state-of-the-arts under arXiv and GovReport. In the next two sections, we will investigate whether the advancement in summarization research has brought us far enough to build a robust summarization system (i.e., model and metric) based on the summaries generated from all of the 24 implemented summarizers in this section. For consistency, we will refer to Longformer-based sparse attention BART and PEGASUS as BART/PEGASUS (LEAD \#K) as it only takes the leading input token, whereas, reduce-then-summarize models will be referred to as BART/PEGASUS (ORACLE \#K). The \# symbol represents the token input length limit of the Transformer-based summarizer. 
\section{Human Evaluation of Models}\label{sec. abs_model}
To assess the overall quality of summaries, we randomly sampled 204 model-generated summaries from each dataset to be evaluated by three annotators based on the relevance and factual consistency aspect. To ensure comparability between model variants, we randomly sampled document IDs from the test set and extracted all 12 corresponding model summaries to annotate. As each summary ranged from 5 to 15 sentences, we annotated 4,190 sentences, matching a large-scale human evaluation by \citet{pagnoni-etal-2021-understanding} of 2,250 short-document articles.

\subsection{Annotation Procedures}
\paragraph{Relevance} Relevance measures whether a summary contains the main ideas of the source. As the author is arguably the best person to summarize the source, we assign relevance scoring based on the percentage of the reference summary's main ideas contained in the generated summary. The relevance score of each summary is the average of three annotation samples. 

\paragraph{Factual Consistency} Factual consistency measures whether a candidate summary is factually consistent with the source. Following \citet{pagnoni-etal-2021-understanding}, we classify each summary sentence's factuality based on seven types of errors: i) \textit{PredE} - predicate in summary inconsistent with source, ii) \textit{EntityE} - primary arguments or its attributes are wrong, iii) \textit{CircE} - predicate's circumstantial information is wrong, iv) \textit{CorefE} - co-reference error, v) \textit{LinkE} - multiple sentences linked incorrectly, vi) \textit{OutE} - out of article error and vii) \textit{GramE} - unreadable sentence(s) due to grammatical errors. Similarly, the factual consistency of a summary is the percentage of factually consistent sentences and the final score is the average of three samples. 

\paragraph{Inter-Annotator Agreement} Following \citet{fabbri2021summeval}, the inter-annotator interval kappa of relevance score between the three annotators is 0.5874, computed based on Krippendorff’s alpha coefficient \cite{krippendorff2011computing} where each score is assigned to a multiple of quarter intervals. To calculate inter-annotator agreement of factual consistency, we follow \citet{durmus2020feqa,pagnoni-etal-2021-understanding} in using Fleiss Kappa, $\kappa$, and the percentage, $p$, of annotators that agree with the majority class. With a total of 4190 sentences, we observe $\kappa = 0.52$ and $p = 84\%$, slightly lower but comparable to \citet{pagnoni-etal-2021-understanding}'s result ($\kappa = 0.58$ and $p = 91\%$).

\subsection{Long Abstractive Model Analysis}

\begin{figure}[!h]
    \centering
       \includegraphics[width=0.48\textwidth]{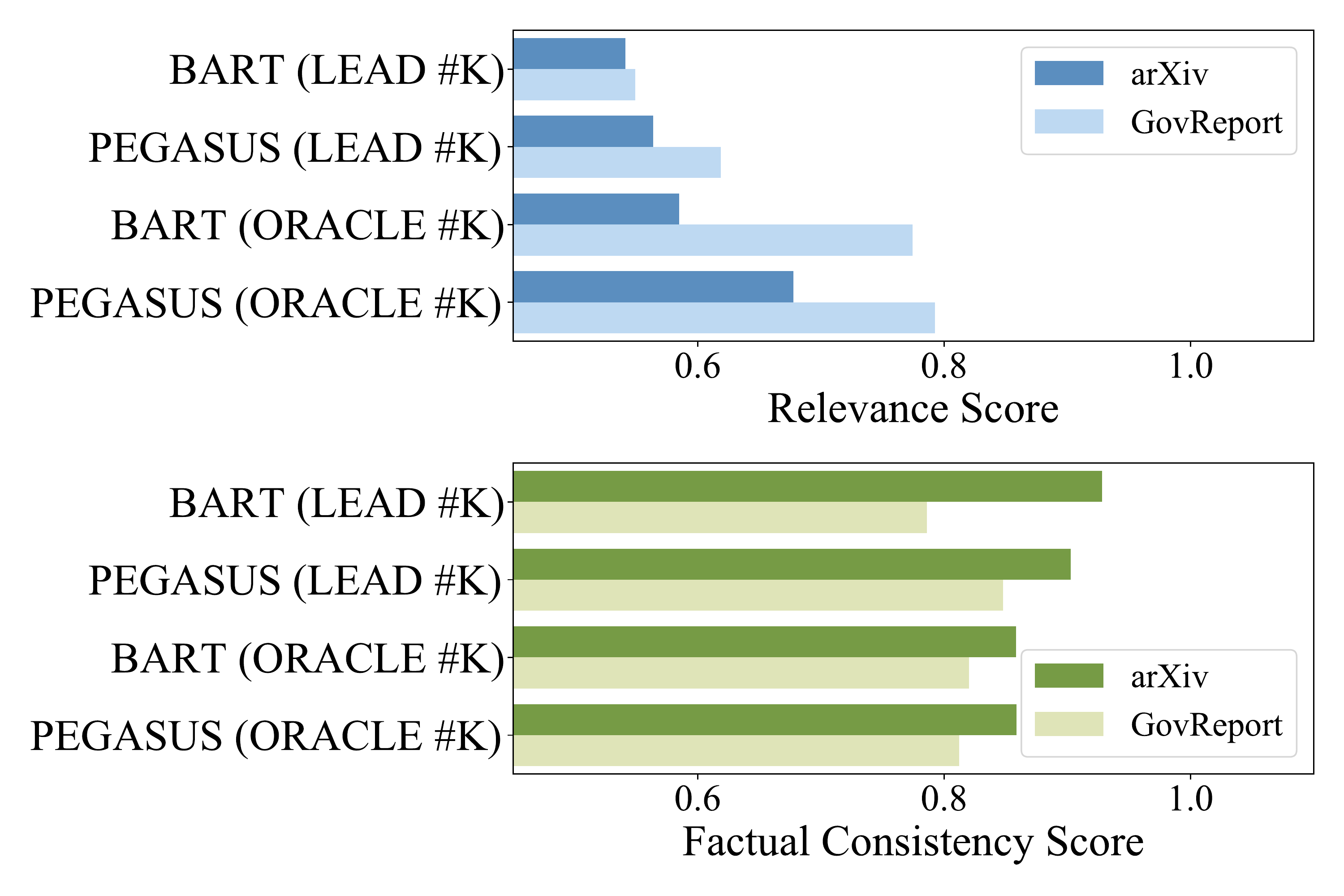}\\
       \vspace{-0.3cm}
       \caption{\textit{Average} human relevance (top) and factual consistency (bottom) scores for BART and PEGASUS models with 1K, 4K and 8K input limit.}
       \label{fig. human_summ_level}
\end{figure}

\paragraph{Relevance} Benefiting from processing the oracle inputs, Figure \ref{fig. human_summ_level} shows BART/PEGASUS (ORACLE \#K) to achieve a higher relevance score than BART/PEGASUS (LEAD \#K). On average, PEGASUS also performs better than BART. Looking at the models with the same pre-training tasks, we observe that BART (ORACLE \#K) did not significantly outperform BART (LEAD \#K) on arXiv. On the other hand, PEGASUS (ORACLE \#K) shows a significant improvement over PEGASUS (LEAD \#K) under both the arXiv and GovReport dataset. We hypothesize that when models take the oracle inputs, the text becomes incoherent and the immediate connection between sentences is less obvious, causing it harder for BART models to understand the contextual dependencies between the tokens. In contrast, PEGASUS's Gap-Sentence Generation pre-training may help models in reasoning the contextual dependencies of an incoherent text.

\paragraph{Factual Consistency} On average, we also observe PEGASUS makes fewer factual errors as compared to BART across most settings. Unlike the relevance aspect, the BART/PEGASUS (ORACLE \#K) setting often achieves lower factual consistency results as compared to BART/PEGASUS (LEAD \#K). This indicates that while models can more easily capture relevant text, incoherent texts may cause them to make more factual errors. As BART/PEGASUS (ORACLE \#K) utilize an oracle retriever during testing that is not allowed under normal settings, similar issues could potentially be exacerbated when a model-based retriever \cite{pilault2020extractive,manakul2021longspan,mao-etal-2022-dyle} is used to extract salient sentences from the source. Finally, this also indicates that maximizing ROUGE itself leads us to models with more relevant summaries but may not be necessarily factual.

\paragraph{Summary Quality v.s. Input Limit} Other than high-level analysis of the different pre-training and mechanism results, we investigate the relationship of the adjustment in input limit of different Transformer variants against the human-annotated relevance and factual consistency scores. 
\begin{table}[!ht]
\resizebox{0.48\textwidth}{!}{%
\begin{tabular}{|l|ll|ll|}
\hline
\multicolumn{1}{|l|}{\multirow{2}{*}{\textbf{Model Configuration}}} & \multicolumn{2}{c|}{\textbf{arXiv}}                             & \multicolumn{2}{c|}{\textbf{GovReport}}                         \\ \cline{2-5} 
\multicolumn{1}{|c|}{}                               & \multicolumn{1}{c}{\textbf{REL}} & \multicolumn{1}{c|}{\textbf{FACT}} & \multicolumn{1}{c}{\textbf{REL}} & \multicolumn{1}{c|}{\textbf{FACT}} \\ \hline
BART \quad\;\;\, (LEAD \#K)                                      &      $+2.78^{**}$                    &            $+1.98^{*}$               &      $+3.03^{**}$                     &           $+1.03$                 \\
PEGASUS (LEAD \#K)                                   &    $+0.31$                    &                $+0.81$           &      $+1.95^{**}$                    &                  $+0.67$          \\
BART \quad\;\;\, (ORACLE \#K)                                    &    $-0.52$                  &        $-0.29$                   &     $-0.61$                     &     $+0.03$                       \\
PEGASUS (ORACLE \#K)                                 &     $+0.13$                    &     $-0.14$                       &    $+0.21$                      &                  $+0.12$          \\ \hline
\end{tabular}}
\vspace{-0.2cm}
\caption{Coefficient of simple linear regression of Relevance and Factual Consistency against Input Limit. Values are in percentage point per 1K input limit. * represents $p < 0.05$ and ** represents $p<0.001$.}
\label{tab. human_eval_model}
\end{table} Table \ref{tab. human_eval_model} shows that the relevance score increases when the input limit of the BART/PEGASUS (LEAD \#K) models is extended but does not show meaningful differences when the oracle input length of the BART/PEGASUS (ORACLE \#K) models is adjusted. Since longer oracle input length increases the difficulty of identifying salient content for a BART/PEGASUS (ORACLE \#K) model and the increase in difficulties did not lead to a drop in summarization performance, this suggests that both pre-trained Transformers are capable of reasoning through long-range texts. This also indicates that the gain in relevance score mostly comes from the reduction in information loss caused by the input limit of Transformer-based summarizers. 

While we see an improvement in factual consistency scores when vanilla pre-trained Transformers increase their input limits using Longformer, only BART (LEAD \#K) under arXiv shows a statistically significant result. The BART/PEGASUS (ORACLE \#K) models do not show conclusive results as to which configurations will generate summaries that are most factually consistent. 

\subsection{Fine-grained Analysis of Factual Errors}
\begin{figure} [!h]
    \centering
       \includegraphics[width=0.48\textwidth]{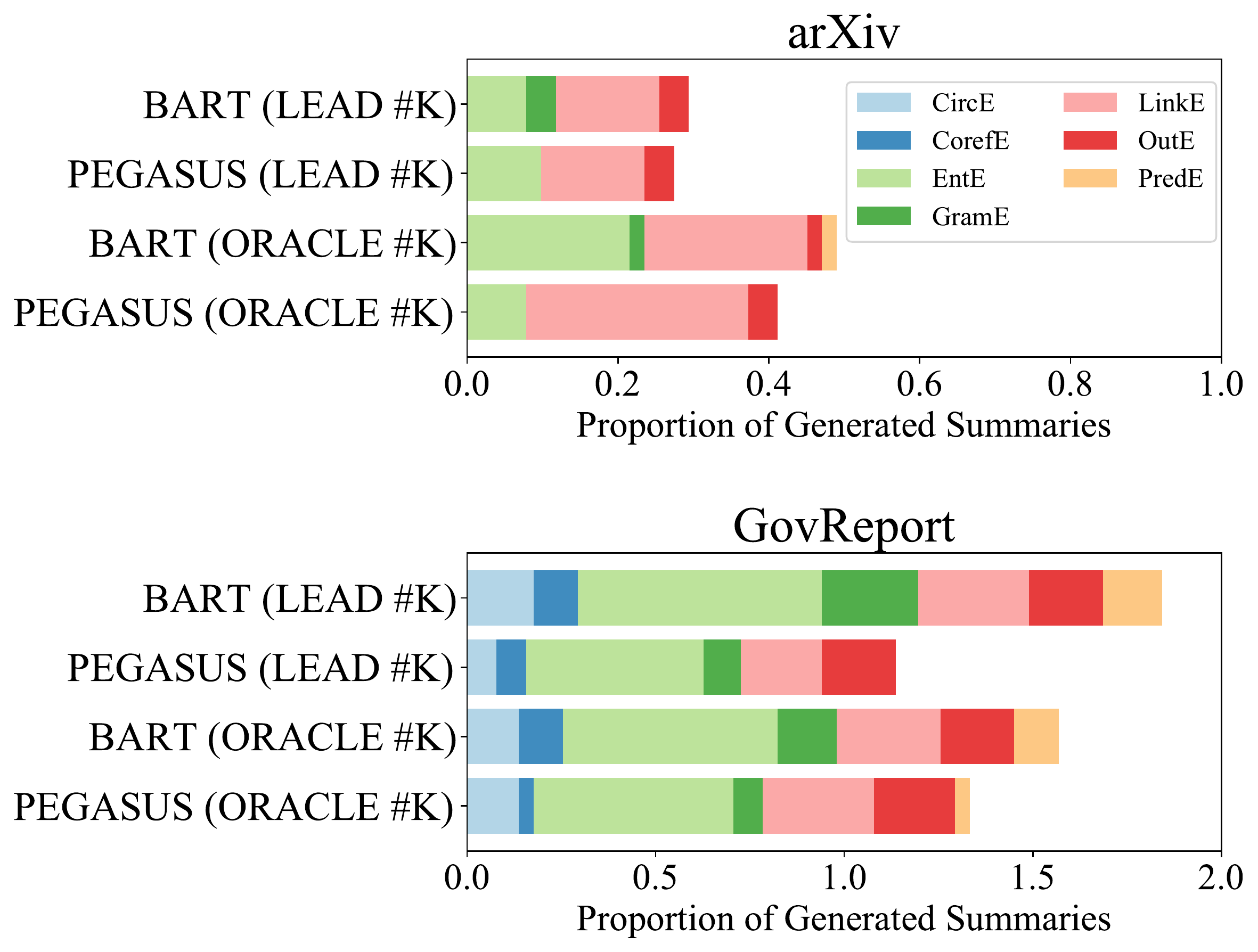}\\
       \vspace{-0.3cm}
       \caption{\textit{Average} Proportion of Factual Error Type for all generated summaries of BART and PEGASUS models with 1K, 4K, and 8K input limits. As a long document summary have multiple sentences and can have multiple error types, the total proportion may exceed 1.} 
       \label{fig. factual}
\end{figure}

Under real-world scenarios, a model will not be evaluated based on the percentage of factual sentences and is only considered robust if it generates summaries that are almost entirely error-free. However, the models generate factually \textit{inconsistent} summaries, on average, 35\% and 81\% of the time under arXiv and GovReport respectively. The least errors are made by PEGASUS (LEAD 8K) in arXiv (21\%) and PEGASUS (ORACLE 1K) in GovReport (60\%). Given the unacceptably high amount of factual errors, it is fair to conclude that the models are not sufficiently robust. Thus, it is more important that we analyze the type of errors they made and how we can improve their performance in the factuality aspect. To this end, we investigate the proportion of summaries with different types of factual error instances in Figure \ref{fig. factual}. 

\begin{table*}[ht]
\resizebox{\textwidth}{!}{%
\begin{tabular}{l|cccccccc|cccccccc}
\thickhline
 & \multicolumn{8}{c|}{\textbf{Relevance}} & \multicolumn{8}{c}{\textbf{Factual Consistency}} \\ 
 & \multicolumn{4}{c}{\textbf{arXiv}} & \multicolumn{4}{c|}{\textbf{GovReport}} & \multicolumn{4}{c}{\textbf{arXiv}} & \multicolumn{4}{c}{\textbf{GovReport}} \\ \thickhline
\multicolumn{1}{l|}{\multirow{2}{*}{\textbf{Metrics}}} & \multicolumn{2}{c}{Pearson} & \multicolumn{2}{c|}{Spearman} & \multicolumn{2}{c}{Pearson} & \multicolumn{2}{c|}{Spearman} & \multicolumn{2}{c}{Pearson} & \multicolumn{2}{c|}{Spearman} & \multicolumn{2}{c}{Pearson} & \multicolumn{2}{c}{Spearman} \\
\multicolumn{1}{c|}{} & $\rho$ & p-val & $r$ & \multicolumn{1}{c|}{p-val} & $\rho$ & p-val & $r$ & p-val & $\rho$ & p-val & $r$ & \multicolumn{1}{c|}{p-val} & $\rho$ & p-val & $r$ & p-val \\ \hline
\textbf{BLEU} & 0.21 & 0.00 & 0.21 & \multicolumn{1}{c|}{0.00} & 0.37 & 0.00 & 0.35 & 0.00 & -0.05 & 0.48 & -0.05 & \multicolumn{1}{c|}{0.45} & -0.12 & 0.09 & -0.14 & 0.11 \\
\textbf{METEOR} & \underline{0.26} & 0.00 & 0.22 & \multicolumn{1}{c|}{0.00} & 0.40 & 0.00 & 0.38 & 0.00 & 0.08 & 0.24 & 0.09 & \multicolumn{1}{c|}{0.18} & -0.09 & 0.14 & -0.13 & 0.12 \\
\textbf{ROUGE-1} & \textbf{0.29} & 0.00 & \textbf{0.25} & \multicolumn{1}{c|}{0.00} & \textbf{0.53} & 0.00 & \textbf{0.52} & 0.00 & -0.08 & 0.26 & -0.13 & \multicolumn{1}{c|}{0.16} & -0.12 & 0.09 & -0.11 & 0.12 \\ 
\textbf{ROUGE-2} & 0.14 & 0.03 & 0.16 & \multicolumn{1}{c|}{0.02} &  \underline{0.43} &  0.00  & \underline{0.44} & 0.00 & -0.12 & 0.09 & -0.13 & \multicolumn{1}{c|}{0.10} & -0.08 & 0.32 & -0.11 & 0.10 \\
\textbf{ROUGE-L} & 0.12 & 0.07 & 0.17 & \multicolumn{1}{c|}{0.06} &  0.38 & 0.00 & 0.39 & 0.00 & -0.16 & 0.09 & -0.15 & \multicolumn{1}{c|}{0.07} & -0.08 & 0.21 & -0.11 & 0.11 \\
\textbf{BERTS} & 0.22 & 0.00 & 0.18 & \multicolumn{1}{c|}{0.00} & 0.38 & 0.00 & 0.38 & 0.00 & -0.09 & 0.12 & -0.10 & \multicolumn{1}{c|}{0.10} & 0.00 & 0.95 & -0.04 & 0.57 \\
\textbf{BARTS-ZS} & 0.06 & 0.44 & 0.12 & \multicolumn{1}{c|}{0.09} & 0.19 & 0.00 & 0.25 & 0.00 & 0.25 & 0.00 & 0.24 & \multicolumn{1}{c|}{0.03} & 0.17 & 0.06 & 0.06 & 0.02 \\
\textbf{BARTS-FT} & 0.00 & 0.98 & 0.03 & \multicolumn{1}{c|}{0.64} & 0.18 & 0.00 & 0.24 & 0.00 & \textbf{0.32} & 0.00 & \textbf{0.36} & \multicolumn{1}{c|}{0.02} & \textbf{0.51} & 0.00 & \textbf{0.48} & 0.00 \\
\hline
\textbf{OpenIE} & 0.21 & 0.00 & \underline{0.23} & \multicolumn{1}{c|}{0.00} & 0.03 & 0.60 & 0.01 & 0.88 & 0.20 & 0.00 & 0.15 & \multicolumn{1}{c|}{0.03} & 0.33 & 0.00 & 0.34 & 0.00 \\
\textbf{MNLI-TE} & 0.03 & 0.72 & 0.03 & \multicolumn{1}{c|}{0.69} & 0.08 & 0.27 & 0.05 & 0.45 & -0.08 & 0.19 & -0.04 & \multicolumn{1}{c|}{0.56} & -0.14 & 0.18 & -0.13 & 0.20 \\
\textbf{FactCC} & 0.13 & 0.07 & 0.13 & \multicolumn{1}{c|}{0.07} & 0.05 & 0.52 & 0.04 & 0.55 & 0.22 & 0.00 & 0.19 & \multicolumn{1}{c|}{0.00} & 0.28 & 0.00 & 0.27 & 0.00 \\ 
\textbf{FEQA} & 0.03 & 0.66 & 0.01 & \multicolumn{1}{c|}{0.83} & 0.09 & 0.20 & 0.10 & 0.14 & 0.06 & 0.36 & 0.05 & \multicolumn{1}{c|}{0.45} & -0.08 & 0.24 & 0.00 & 0.46 \\
\textbf{QUAL} & -0.06 & 0.38 & -0.07 & \multicolumn{1}{c|}{0.34} & 0.30 & 0.00 & 0.34 & 0.00 & 0.12 & 0.07 & 0.16 & \multicolumn{1}{c|}{0.02} & 0.12 & 0.08 & 0.10 & 0.11 \\ 
\textbf{SummaC} & 0.09 & 0.22 & 0.08 & \multicolumn{1}{c|}{0.24} & 0.05 & 0.49 & 0.04 & 0.57 & \textbf{0.32} & 0.00 & \underline{0.32} & \multicolumn{1}{c|}{0.00} & \underline{0.39} & 0.00 & \underline{0.38} & 0.00 \\  \thickhline
\end{tabular}%
\caption{Statistical Relationship between human judgment (relevance and factual consistency) and metric scores based on Pearson correlation, $\rho$, and Spearman rank correlation, $r$, coefficients and their p-values. Upper and lower part show results for general metric and factual consistency metric respectively.}
\label{tab. metric}}
%\vspace{-0.3cm}
\end{table*}

As arXiv articles are pre-processed when the dataset was introduced by \citet{cohan2018discourse} while GovReport articles closely resemble the original documents \cite{huang2021efficient}, the task is made less challenging under arXiv, and mistakes related to CorefE, EntE and CircE are greatly reduced. Still, models under the arXiv setting generate higher LinkE errors as they are required to paraphrase the source text more. We also see BART (ORACLE \#K) and PEGASUS (ORACLE \#K) to make more LinkE errors as the oracle input text is less coherent as compared to the leading input text. We again observe PEGASUS makes fewer errors as compared to BART. The better performance of PEGASUS mostly comes from making fewer CorefE, EntE, GramE and PredE errors. 

We conclude this section by noting that while ROUGE scores show minor differences between BART and PEGASUS, human evaluation relevance and factual consistency scores reveal that PEGASUS is considerably better than BART. This conflicts with the findings of \citet{rothe2021thorough} that PEGASUS task-specific pre-training did not bring improvement in summarization performances, emphasizing the need of evaluating summaries based on the quality judged by a summary user rather than solely relying on the ROUGE metric. 

\subsection{Dataset for Metrics Evaluation} We release the human-annotated summaries to encourage the future exploration of long document models and metrics\footnote{https://github.com/huankoh/How-Far-are-We-from-Robust-Long-Abstractive-Summarization}. In the next section, we utilize the dataset to assess long document metrics.

\section{Human Evaluation of Metrics}\label{sec. eval_metric}
With high factual inconsistency rates, long abstractive summarizers remained unready for real-world implementation. It is thus paramount to ensure that the performances of future proposed models can be evaluated by metrics that are well correlated with user judgment. However, previous works on evaluation metrics have mainly focused on short document summarization research settings due to (i) the lack of human-annotated long document model-generated summaries and (ii) the reliance of metrics on pre-trained language models that are fine-tuned on short document datasets \cite{maynez2020faithfulness,durmus2020feqa,wang2020asking}. Relying on our annotated dataset, we adapt evaluation metrics proposed in prior works to the long document settings and correlate their metric scores with average human relevance and factual consistency scores.

\paragraph{General Metric}
General metrics attempt to capture the overall summary quality including relevance and factual consistency. Assessed general metrics are: BLEU \cite{papineni2002bleu}, ROUGE \cite{lin2004rouge}, METEOR \cite{banerjee2005meteor}, BERTScore \cite{zhang2019bertscore} and BARTScore \cite{yuan2021bartscore}. We implement zero-shot (BARTS-ZS) and fine-tuned (BARTS-FT) BARTScore. BART-ZS uses the original BART model while BART-FT is fine-tuned on the arXiv and GovReport datasets. Both are extended to 8K tokens using Longformer.

\paragraph{Factual Consistency} Factual consistency metrics we assess are: OpenIE \cite{goodrich2019assessing} that extracts semantic triples from source and summary, then compute scores through embedding matching \cite{reimers2019sentence}. FactCC \cite{kryscinski-etal-2020-evaluating} adopts a weakly-supervised model approach. FEQA \cite{durmus2020feqa} and QUAL \cite{nan2021improving} evaluate factuality using a question-generation and answering (QGA) approach. TE-MNLI \cite{maynez2020faithfulness} and SummaC \cite{laban2022summac} are text entailment approach, TE-MNLI evaluates probability of entailment at the document-level while SummaC at the sentence-level. For metrics with short input limits, we extend the input limit of FactCC using Longformer and use the oracle summaries as a substitute for the source for FEQA, QUAL and TE-MNLI. Implementation details in Appendix \ref{subsec:fact_implementation}. 

\subsection{Overall Result} 
\paragraph{Relevance} Contrary to past research under short-document setting \cite{kryscinski2019neural,bhandari2020re,akter2022revisiting}, Table \ref{tab. metric} shows that ROUGE scores still correlate best with the human judgment of relevance score in our settings. This provides comfort for future research to rely on the ROUGE metric for benchmarking long document abstractive models in generating relevant summaries. We hypothesize that the effectiveness of ROUGE metric is due to the linguistic styling of long document datasets that are often written in formal languages. We caution that similar results may not be achieved by ROUGE metric when the dataset and model-generated summaries are sufficiently abstractive.

\paragraph{Factual Consistency} The metrics that achieve the best overall correlation with the human factual consistency scores are fine-tuned BARTScore, followed by SummaC, FactCC, and OpenIE. Interestingly, zero-shot BARTScore also achieves third and fifth-best results on arXiv and GovReport respectively. Consistent with \citet{pagnoni-etal-2021-understanding}, QGA approaches do not seem to achieve statistically significant results, except for QUAL under GovReport. From the perspective of efficiencies, BARTScore and FactCC require approximately 4 days of fine-tuning per dataset on an RTX 3090 GPU while zero-shot SummaC and OpenIE can be implemented immediately without dataset-specific training. On balance, SummaC and BARTS-FT seem to stand out from the rest as the most effective zero-shot and fine-tuned metric respectively. Nevertheless, it is more important to thoroughly investigate why and when the metrics will identify factual inconsistencies in model outputs. 

\subsection{Identification of Factual Error Types} 
\begin{figure} [!h]
    \centering
       \includegraphics[width=0.48\textwidth]{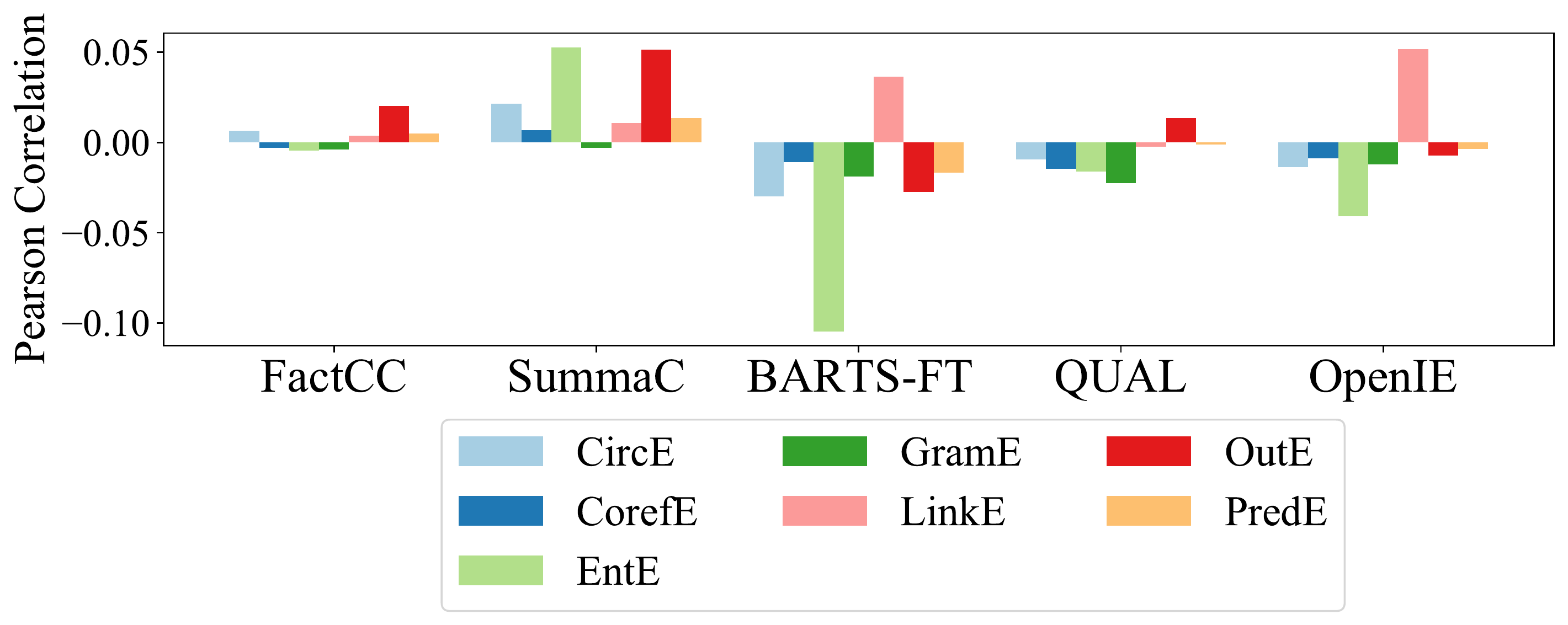}\\
       \vspace{-0.25cm}
       \caption{Change in Pearson correlation when error types are omitted. Higher value indicates a greater influence of the error type on overall correlation result.}
       \label{fig. factual_type}
\end{figure}

Overall correlation with human factual consistency score does not reveal the limitations of a metric in identifying different types of factual errors \cite{goyal2021annotating,pagnoni-etal-2021-understanding}. Hence, we plot the contribution of each error type to the overall correlation in Figure \ref{fig. factual_type}. It shows the change in correlation when the error type is excluded from the calculation. As compared to Table \ref{tab. metric}, a higher positive bar value shows that the error type contributed more to the metric performances, causing a decrease in overall correlation.

Figure \ref{fig. factual_type} shows that OpenIE and BARTScore are not able to identify entity errors (EntE) well. We hypothesize that this is because OpenIE relies on the soft-embedding similarity while BARTScore finds reasonableness in generating closely related entities in the source document. Nevertheless, BARTScore and OpenIE show better ability at identifying sentence linkage (LinkE) errors as BARTScore takes the full context of the entire generated summary into account while OpenIE assesses the relationship between semantic triples. FactCC, SummaC and QUAL which only relied on sentence- or question-level granularity did not see a high correlation with LinkE as they do not take the overall contexts of the generated summaries. SummaC shows strong correlations with entity (EntE) and out-of-article (OutE) errors. As different metrics can better identify different factual error types, combining the advantages of various metrics to address their limitations may be worthwhile. For a simple illustration, by taking the average normalized metric scores of BARTS-FT and SummaC, we are able to increase Table \ref{tab. metric}'s best Pearson correlation result of arXiv from 32\% to 38\% and GovReport from 51\% to 59\%, representing an absolute percentage point increase of 6\% and 8\% respectively. 

\subsection{On the Effectiveness of BARTScore} Given the superiority of BARTScore as a factuality metric, we further analyze it in detail. BARTScore relies on a BART's average log-likelihood of generating the evaluated summary conditional on the source document: $\frac{1}{m}\textstyle{\sum^m_{t=1}} \log p(\mathbf{y_t}|\mathbf{y_{<t}},\mathbf{d})$ where $\mathbf{y_t}$ represent generated tokens in the summary at generation step $t$ while $\mathbf{d}$ represents source \cite{yuan2021bartscore}. Under the fine-tuned variant, BARTScore is fine-tuned as a summarization model. Thus, a lower BARTScore indicates that the BART model shows a lower likelihood of generating the evaluated texts. This suggests that summarization models are "aware" of potentially making factual errors in the form of lower generation probability. Similar to our findings, \citet{xu2020understanding} has found that lower generation probability (and higher entropy value) leads to greater novelty in the tokens generated but a higher chance of factual inconsistencies under short-document settings. Consequently, solving the factuality aspects of abstractive models and metrics from this perspective may be a fruitful direction to explore. 

In addition, we fine-tuned BARTScore on different datasets and compute its correlation with human factual consistency scores in Table \ref{tab. fine_tuning_metrics}. BART shows a better correlation when metrics are fine-tuned on in-domain datasets. In particular, we find the best results are achieved for arXiv when BART is fine-tuned on arXiv or PubMed and for GovReport when BART is fine-tuned on GovReport.

\begin{table}[!ht]
\resizebox{0.48\textwidth}{!}{%
\begin{tabular}{|l|cc|cc|}
\hline
\multicolumn{1}{|c|}{\multirow{2}{*}{\textbf{Variants}}} & \multicolumn{2}{c|}{\textbf{arXiv}}                             & \multicolumn{2}{c|}{\textbf{GovReport}}                         \\ \cline{2-5} 
\multicolumn{1}{|c|}{}                               & \multicolumn{1}{c}{Pearson} & \multicolumn{1}{c|}{Spearman} & \multicolumn{1}{c}{Pearson} & \multicolumn{1}{c|}{Spearman} \\ \hline
Zero-Shot                                       &      $0.25$                   &            $0.24$                &      $0.23$                  &           $0.17$                 \\
arXiv                     &    $0.32$                   &                $0.36$           &          $0.19$           &                  $0.18$          \\
GovReport                                    &    $0.31$                  &        $0.21$                    &      $\mathbf{0.51}$                    &    $\mathbf{0.48}$                       \\
%CNN-DM      &     $0.00$                    &     $0.02$                       &     $0.12$                      &                  $0.02$         \\
PubMed                                    &    $\mathbf{0.34}$                  &      $\mathbf{0.36}$                 &      $0.38$                    &     $0.38$                       \\
BookSum                                    &    $0.31$                  &       $0.31$                    &      $0.36$                     &     $0.31$                      \\
 \hline
\end{tabular}}
\caption{Human Factual Consistency correlation with BARTScore variants fine-tuned on different datasets. All results are statistically significant, where $p < 0.05$.}
\label{tab. fine_tuning_metrics}
\vspace{-0.25cm}
\end{table} 
To validate this hypothesis, we further implement FEQA with Sci-BERT \cite{beltagy2019scibert} fine-tuned on SQuaD \cite{rajpurkar2016squad,rajpurkar2018know} and QUAC \cite{choi2018quac} and we obtain statistically significant Pearson correlation ($\rho=+0.22$) on arXiv, a four-fold increase as compared to the original variant. This finding strongly emphasizes the importance of fine-tuning metrics on in-domain datasets. Future work on metrics could thus benefit from incorporating fine-tuning strategies \cite{kryscinski-etal-2020-evaluating,laban2022summac} rather than relying merely on publicly available models \cite{maynez2020faithfulness,durmus2020feqa}. Importantly, the fine-tuning strategy should be efficient and generalizable to other domains to ensure that it is not limited to short news articles. 
\section{Conclusion} 
In this work, we perform human evaluations of model-generated summaries to critically analyze the relevance and factual consistency aspect of models and metrics under long document settings. 

For models, we highlight that the constant strive for higher ROUGE scores leads us to long document models with more relevant summaries but not necessarily factual ones. We also show that PEGASUS pre-training allows long document Transformer to make fewer factual errors and can comprehend incoherent text better, suggesting that PEGASUS can be more beneficial than BART for reduce-then-summarize architectures that are common for long document summarizers. For metrics, we observe that ROUGE remains superior at assessing the relevance of summaries, while a fine-tuned BARTScore can be most effective in evaluating the factuality of long document summaries. 

We also release the annotated dataset to encourage analysis of summarization systems across a broader range of settings. We hope that this work can provide practical insights into future research to develop long document summarization systems that can be relied upon in our daily lives. 
\section*{Limitations}
Our findings and conclusions relied on human annotation efforts by three annotators. To balance the quality and quantity of annotation, three annotators evaluated the same 408 summary-document pairs across two datasets. While having three annotations per summary-document pair reduces the variability and enhances the final quality of annotation, increasing the size and diversity of our annotated dataset would further enhance the statistical significance of our findings. 

Prior works on summarization metrics have assessed their performances on short summary-document pairs and often relied on pre-trained models with token limits that cannot be easily extended. While we have taken reasonable steps in adapting their methods to long document settings, it is plausible that better adaptation approaches can be discovered.

Finally, our experiments are conducted on the arXiv and GovReport benchmark datasets. The documents in both datasets are written in formal language. While formal language is common across long document benchmark datasets, this may result in domain bias. Our experimental processes and findings may also be limited to the English language. This is especially the case for our human-annotation process as we relied on English grammatical rules to determine the qualitative aspects of model-generated summaries. Thus, our processes and findings are likely not applicable to long documents that are not written in English. Nevertheless, we hope that our work can indirectly inspire or be extended to the research in multilingual long document summarization.

% Entries for the entire Anthology, followed by custom entries
\bibliography{anthology,custom}
\bibliographystyle{acl_natbib}
\newpage
\appendix
\section{Appendices}
\subsection{Broader Impacts}
Abstractive models implemented are in general neural conditional generation models that have a wide range of capabilities due to their ability to carry out arbitrary language generation tasks. This may have a negative society such as generating texts that are biased towards certain minorities or unfairly discriminate against a certain group. This risk may, for example, arise from the human-annotated model dataset that we aim to release along with this work. Nevertheless, we have taken sufficient care to ensure that the potential risks of broad negative impacts are minimized. Based on our annotation, we believe that the risks of negative broader impacts are well manageable. 

\subsection{Human Annotated Dataset}
All of the 408 human-annotated summaries are randomly sampled from the summaries generated from our implemented models on arXiv and GovReport dataset. To ensure that our model summaries are annotated by human experts, we recruited three volunteers. One has years of industry experience in accounting and finance with CIMA certification while the other two are Ph.D. students of public health and computer science. Our aim for the release of the human-annotated dataset is to encourage the development of a factual consistency summarization system (model and metric). The dataset is intended for research use only. Other than that of what is already publicly available, we have taken extra steps to ensure that the factual inconsistencies generated by the summarization models do not discriminate against any individual or uniquely identify a certain person, thereby leaking information.

\subsection{Model Implementation}\label{subsec:model_implementation}
Our model experiment in section 3 was implemented on the arXiv and GovReport with train/validation/test split of 203,037/6,436/6,440 and 17,519/974/973 respectively. Given two different pre-trained Transformers with three different input limit lengths that were tested on the baseline Longformer-only BART/PEGASUS models as well as upper-bound reduce-then-summarize models. This gives us twelve model variations per dataset. For 1K token configurations, we use BART-large and PEGASUS-large. For 4K and 8K token configurations, we follow Longformer's implementation in extending the position embedding to 4K and 8K tokens by repeatedly copying BART-large and PEGASUS-large's 1K position embeddings multiple times. All models are trained with teacher forcing on the same RTX 3090 GPU with 24 GiB of GPU memory. To save memory, we implemented gradient checkpoint. For all models with have an effective batch size of 16 where the batch size is set to be 2 and gradient accumulation step set to 8. The most expensive experiments of 8K limit require approximately 3 and 4 days respectively for Longformer-BART and Longformer-PEGASUS. As ROUGE tends to prefer longer summaries \cite{sun2019compare}, we fix the maximum model output length to be 512 tokens. Generation parameters of beam search is 5 and length penalty is set to 2.0. %Finally, We use rouge-score package for ROUGE metric. 

\subsection{Factual Consistency Metric Implementation}\label{subsec:fact_implementation}
FEQA, QUAL, FactCC, and TE-MNLI were proposed to evaluate the factual consistency of model-generated summaries under short document settings. They relied on pre-trained Transformer-based models where the input limit of 1024 tokens or lower. To extend these metric models to the long document domain, we adopt two approaches: if (i) the model requires data specific fine-tuning like FactCC, we extend the input limit of the metric model using Longformer, or (ii) the model relies on a pre-trained model that is fine-tuned on other datasets, we extract the oracle summaries of the source document where the length is the input limit of the pre-trained model.

\paragraph{FactCC} FactCC \cite{kryscinski-etal-2020-evaluating} implements a BERT-based factual consistency classifier that is trained on synthetic data, where the positive data labels are non-paraphrased and paraphrased sentences from the source document, and the negative labels are artificially corrupted sentences from the source document. The starting point of the BERT model is uncased, base BERT model pre-trained on English data with 512 token limits. We extend this model to 8,192 tokens using Longformer's implementation. Then, we follow the original author's work in generating the synthetic data to train our extended BERT classifier on RTX 3090 GPU with 24 GiB of GPU memory. 

\paragraph{TE-MNLI} TE-MNLI \cite{maynez2020faithfulness} is a BERT-large classifier fine-tuned on the Multi-NLI
dataset \cite{williams2018broad}. The classifier judged if a summary entails the document, is neutral to the document, or contradicts the document. Multi-NLI is a sentence-level classifier. We tokenize the candidate summary into sentences and separately evaluate the factual consistency of each sentence. The score for a candidate summary equals 1 minus the average probability of contradiction for all sentences in the candidate summary. To adapt the Multi-NLI BERT-large classifier on the long document domain, we limit the total length of summary sentence and document to be less than 512 token lengths by replacing the source document with its oracle summary.  

\paragraph{FEQA and QUAL} FEQA \cite{durmus2020feqa} and QUAL \cite{nan2021improving} measures factual consistency of summaries using a question-generation and question-answering (QGA) approach. This approach employs a question-generation model to generate questions from a given summary output. The generated questions are then measured in two different ways: i) answering the question conditioning on the source and ii) answering the question conditioning on the summary. If the answers match between the source and the summary, the answer is then considered consistent, otherwise, it is inconsistent. QUAL attempts to improve the efficiency of such an approach by combining the question-generation and question-answering steps into a single model. We limit source and candidate summary length to less than 512 tokens by replacing the source document with its oracle summary.

\subsection{Benchmark Dataset Comparison}\label{dataset_comparison}
Long document benchmark datasets studied in this work have been used in prior research to test and compare long document summarization models. arXiv and PubMed \cite{cohan2018discourse} are scientific long document summarization datasets. BigPatent \cite{sharma2019bigpatent} is collected from U.S. patent documents. BillSum is a dataset on summarizing state bills \cite{kornilova2019billsum}. GovReport is a dataset of U.S. Government Accountability Office reports \cite{huang2021efficient}. We also compute the average result of short document datasets based on CNN-DM \cite{nallapati2016abstractive}, NWS \cite{grusky2018newsroom}, XSUM \cite{narayan2018don}, Reddit-TIFU \cite{kim2019abstractive}, and WikiHow \cite{Koupaee2018WikiHowAL} in Table \ref{tab. long_data}. We evaluate the document (D) and summary (S) pairs of benchmark datasets by their compression ratio, extractive coverage, extractive density and uniformity. 

\textit{Compression Ratio} measures the ratio of a source document length against its reference summary length. A higher compression ratio indicates larger information loss in the original document after being summarized. Compression ratios are measured based on tokens and sentences:

$$COMPRESSION_{token} = \frac{|D|}{|S|} $$ $$ COMPRESSION_{sent} = \frac{||D||}{||S||}$$

\textit{Extractive Coverage and Extractive Density} are introduced by \citet{grusky2018newsroom} based on the notion of matching fragments. Fragments are obtained by greedily matching the longest shared token sequence where $\mathcal{F}(D,S)$ reflects a set of fragments with each fragment having a length represented by $|f|$. Extractive coverage calculates the percentage of tokens in summary that is a derivation of the original source text, whereas, extractive density relates to the average squared length of the extractive fragments in the summary. The former indicates the need for a model to coin novel tokens that are not in the original source text while the latter measures whether a model can match the ground truth summary merely by extracting from the original source text without rearranging or paraphrasing text. 
$$COVERAGE(D,S) =  \frac{1}{|S|} \sum_{f \in \mathcal{F}(D,S)} |f| $$
$$DENSITY(D,S) = \frac{1}{|S|} \sum_{f \in \mathcal{F}(D,S)} |f|^2 $$

\textit{Uniformity} measures whether content that are considered important by the reference summary are uniformly scattered across the entire source document. A higher score indicates that important content are scattered across the entire document with no obvious layout bias to take advantage of. This is calculated based on the normalized entropy of the decile positions of salient unigrams in the source text, where salient unigrams are the top 20 keywords extracted\footnote{We use NLTK-RAKE for keywords extraction.}, excluding stopwords, from the reference summary.  
$$UNF(unigram_{pos})= H_{norm}(unigram_{pos}) $$  

\begin{table*}[!ht]
\centering
\resizebox{\textwidth}{!}{%
\small
\begin{tabular}{c|ccccc|ccccc|c}
\thickhline
\multicolumn{1}{l|}{} & \multicolumn{5}{c|}{\textbf{Short Document Datasets}} & \multicolumn{5}{c|}{\textbf{Long Document Datasets}} & {\textbf{Long vs. Short}}\\ [0.5ex]
\multicolumn{1}{l|}{} & CNN-DM & NWS & XSum & WikiHow & Reddit & ArXiv & PubMed & BigPatent & BillSum & GovReport & Avg. Ratio \\ [0.3ex]
\hline
\textbf{\# doc-summ.} & 278K & 955K & 203K & 231K & 120K & 215K & 133K & 1.34M & 21.3K & 19.5K & - \\
\textbf{summ tokens} & 55 & 31 & 24 & 70 & 23 & 242 & 208 & 117 & 243 & 607 & 6.9x \\
\textbf{doc tokens} & 774 & 767 & 438 & 501 & 444 & 6446 & 3143 & 3573 & 1686 & 9409 & 8.3x \\
\textbf{summ sents} & 3.8 & 1.5 & 1 & 5.3 & 1.4 & 6.3 & 7.1 & 3.6 & 7.1 & 21.4 & 3.7x \\
\textbf{doc sents} & 29 & 31 & 19 & 27 & 22 & 251 & 102 & 143 & 42 & 300 & 6.5x \\ [0.3ex]
\hline
\textbf{Compression\textsubscript{token}} & 14.8 & 31.7  & 19.7 & 7.2 & 18.4 & 41.2 & 16.6 & 36.3 & 12.2 & 18.7 &  1.4x \\
\textbf{Compression\textsubscript{sent}} & 8.3 & 22.4 & 18.9 & 3.3 & 14.5 & 44.3 & 15.6 & 58.7 & 9.7 & 18.1 & 2.2x \\
\textbf{Coverage} & 0.890 & 0.855 & 0.675 & 0.610 & 0.728 & 0.920 & 0.893 & 0.861 & 0.913 & 0.942 & 1.2x \\
\textbf{Density} & 3.6 & 9.8 & 1.1 & 1.1 & 1.4 &  3.7 & 5.6 & 2.1 & 6.6 & 7.7 & 1.5x \\
%\textbf{Redundancy} & 0.157 & 0.088 & - & 0.324 & 0.078 & 0.144 & 0.146 & 0.223 & 0.163 & 0.124 & 1.0x \\
\textbf{Uniformity} & 0.856 & 0.781 & 0.841 & 0.813 & 0.777 & 0.894  & 0.896 & 0.922  &  0.903 & 0.932 & 1.2x \\
\thickhline
\end{tabular}%
}
\caption{Comparison of Short and Long Document Summarization Datasets. Intrinsic characteristics are computed based on the average result of test samples. Average Ratios are computed based on the average long over short document statistics.}
\label{tab. long_data}
\end{table*} 

\paragraph{Fundamentals of Long Document} From Table \ref{tab. long_data}, the long document datasets differ from the short documents datasets in two important aspects: document length and compression ratio. Not only that long document datasets have an average document length that is 8.3 times longer than the short document datasets, they also have a considerably higher compression ratio. As compared to short documents, this suggests that either (i) there is a greater compression in the summaries, and/or (ii) the source document contains significantly more redundant information. Both aspects significantly challenge a model's ability to summarize a long document as it is required to reason over long-range dependencies.

\paragraph{Extractiveness and its Relationship with Compression Ratio} 
Looking at the density value, BigPatent and arXiv are significantly less extractive than Pubmed, BillSum and GovReport. Thus, a summarizer is required to have a greater ability at paraphrasing the original document under BigPatent and arXiv. This finding is important as past work in analyzing abstractive summarization of short documents has found that the quality of model-generated summaries \cite{tejaswin2021well,wilber-etal-2021-point} and effectiveness of evaluation metrics \cite{gabriel-etal-2021-go,pagnoni-etal-2021-understanding} to vary based on the extractiveness of benchmark datasets. Intriguingly, we further observe a strongly negative correlation, $\rho = -0.9186$, between the extractive density and the compression ratio metrics. We hypothesize that this is because, under a scenario where summary length is extremely limited, the summary writers are forced to intelligently paraphrase the source concisely so that the reference summaries can cover the salient contents. 

Based on the findings above, we choose GovReport as it is the most extractive dataset with an average compression ratio, and arXiv as it is the second most abstractive dataset with the greatest compression ratio in terms of token for our systematic analysis of long document summarization systems (i.e., models and metrics).

\subsection{Human Evaluation Results for Each Model Variant}
\begin{figure}
    \centering
       \includegraphics[width=0.48\textwidth]{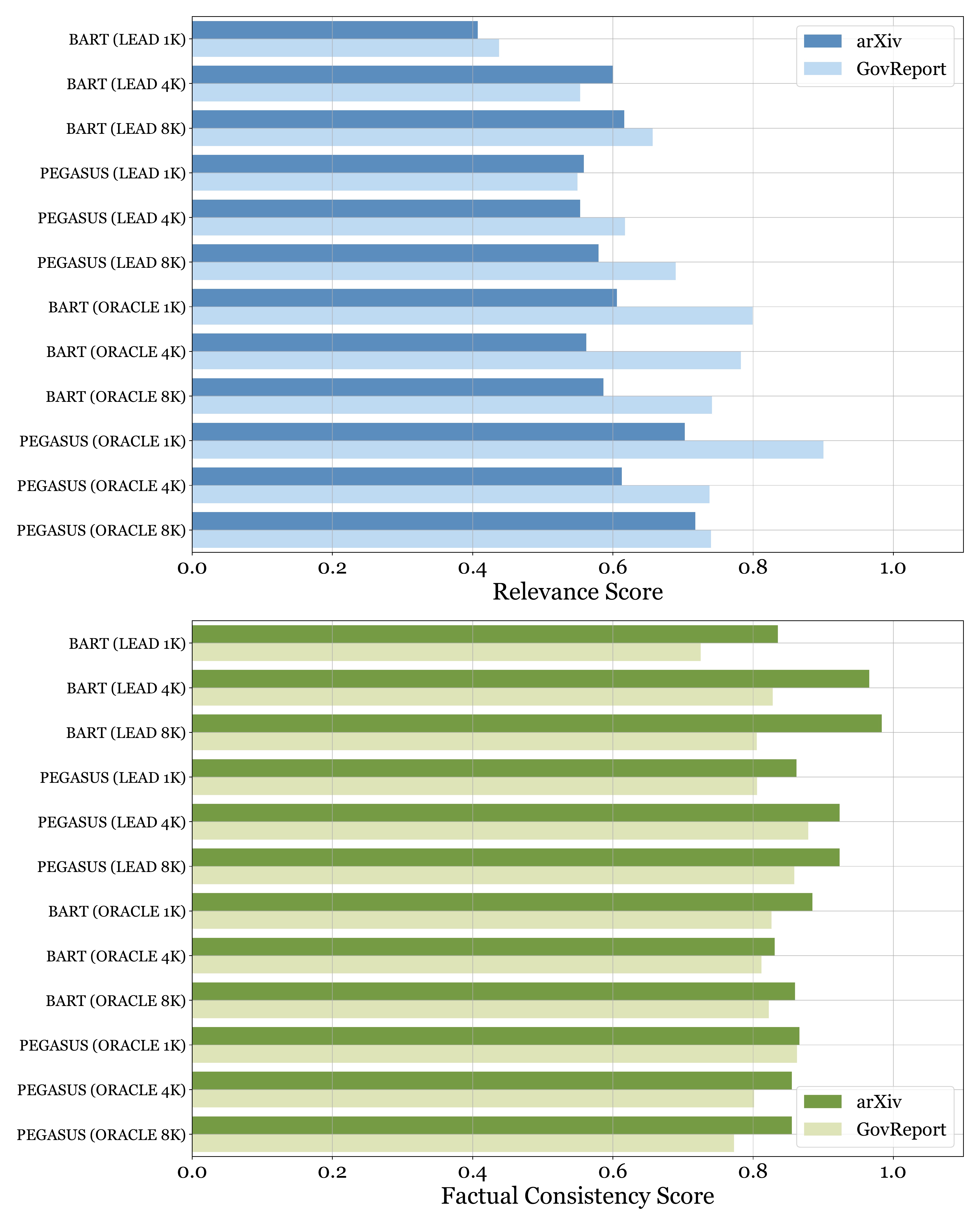}\\
       \caption{Human relevance (upper) and factual consistency scores for each BART model variant.}
       \label{fig. detail_human_summ_level}
\end{figure}

Figure \ref{fig. detail_human_summ_level} shows human evaluation results for each model variant made in the arXiv and GovReport datasets as annotated by our volunteers.

\subsection{Fine-grained analysis of Abstractive Summarizer's Factual Consistency}
\begin{figure}
    \centering
       \includegraphics[width=0.48\textwidth]{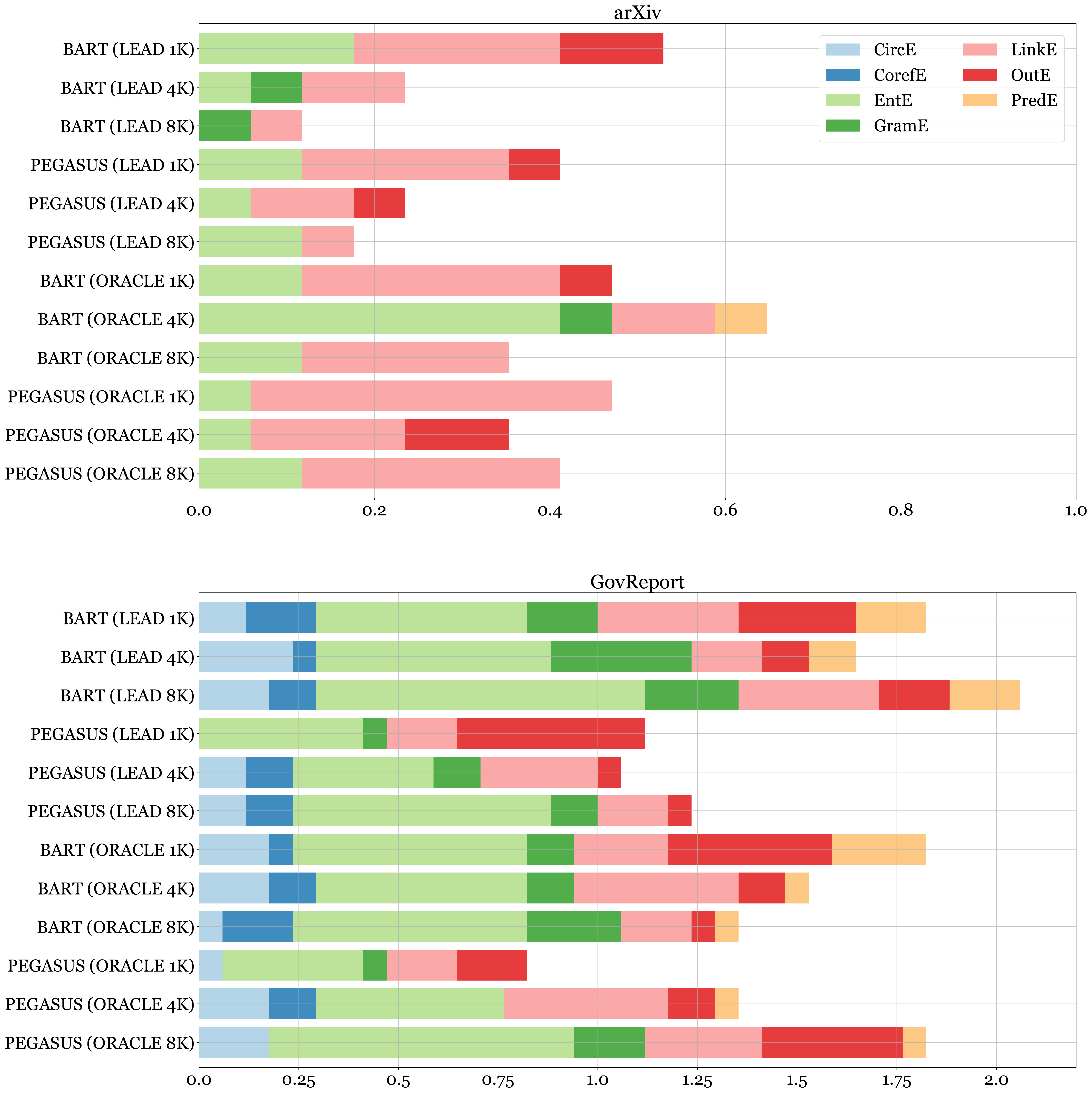}\\
       \caption{Factual Consistency across different model variants. The proportion for each type of error is shown based on the percentage of summaries with the same type of error. As long document summaries may have multiple sentences, each summary may have more than one type of error.}
       \label{fig. detail_factual}
\end{figure}

Figure \ref{fig. detail_factual} shows the types of factual errors that the abstractive models made in the arXiv and GovReport datasets as annotated by our volunteers. As a long document summary have multiple sentences and can have multiple types of errors, the total proportion may exceed 1 but the proportion of errors for each type should be lower than 1.  

\subsection{Human Correlation Results for Precision, Recall, F1 of ROUGE and BERTScore}
Table \ref{tab. rouge_bert_scores} shows the correlation of ROUGE and BERTScore for precision, recall and F1 scores. When measuring relevancy of model-generated summaries, we observe F1 score to often best correlate with human judgment scores. However, when it comes to factual consistency of summaries, we do not see conclusive results as to which variant provide the best results. Furthermore, most results are not statistically significant when measuring factual consistency. Consequently, we do not include these results in our main section.

\begin{table*}[]
\resizebox{\textwidth}{!}{%
\begin{tabular}{l|cccccccc|cccccccc}
\thickhline
 & \multicolumn{8}{c|}{\textbf{Relevance}} & \multicolumn{8}{c}{\textbf{Factual Consistency}} \\ 
 & \multicolumn{4}{c}{\textbf{arXiv}} & \multicolumn{4}{c|}{\textbf{GovReport}} & \multicolumn{4}{c}{\textbf{arXiv}} & \multicolumn{4}{c}{\textbf{GovReport}} \\ \thickhline
\multicolumn{1}{l|}{\multirow{2}{*}{\textbf{Metrics}}} & \multicolumn{2}{c}{Pearson} & \multicolumn{2}{c|}{Spearman} & \multicolumn{2}{c}{Pearson} & \multicolumn{2}{c|}{Spearman} & \multicolumn{2}{c}{Pearson} & \multicolumn{2}{c|}{Spearman} & \multicolumn{2}{c}{Pearson} & \multicolumn{2}{c}{Spearman} \\
\multicolumn{1}{c|}{} & $\rho$ & p-val & $r$ & \multicolumn{1}{c|}{p-val} & $\rho$ & p-val & $r$ & p-val & $\rho$ & p-val & $r$ & \multicolumn{1}{c|}{p-val} & $\rho$ & p-val & $r$ & p-val \\ \hline
\textbf{\underline{ROUGE-1}} & & & & \multicolumn{1}{c|}{} & & & & & & & & \multicolumn{1}{c|}{} &  & & & \\
Precision & 0.08 & 0.26 & 0.07 & \multicolumn{1}{c|}{0.30} & 0.26 & 0.00 & 0.30 & 0.00 & -0.12 & 0.09 & -0.13  & \multicolumn{1}{c|}{0.06} & \textbf{0.12} & 0.10 & \textbf{0.07} & 0.37 \\
Recall & 0.24 & 0.00 & 0.24 & \multicolumn{1}{c|}{0.00} & 0.39 & 0.00 & 0.39 & 0.00 & \textbf{0.18} & 0.00 & \textbf{0.12} & \multicolumn{1}{c|}{0.07} & -0.11 & 0.21 & -0.15 & 0.12 \\
F1 & \textbf{0.29} & 0.00 & \textbf{0.25} & \multicolumn{1}{c|}{0.00} & \textbf{0.53} & 0.00 & \textbf{0.52} & 0.00 & -0.08 & 0.26 & -0.13 & \multicolumn{1}{c|}{0.16} & -0.12 & 0.09 & -0.11 & 0.12 \\ \hline
\textbf{\underline{ROUGE-2}} & & & & \multicolumn{1}{c|}{} & & & & & & & & \multicolumn{1}{c|}{} &  & & & \\
Precision & 0.01 & 0.92 & 0.02 & \multicolumn{1}{c|}{0.83} &  0.27 & 0.00 & 0.36 & 0.00 & -0.16 & 0.03 & -0.21 & \multicolumn{1}{c|}{0.00} & \textbf{0.06} & 0.44 & \textbf{0.00} & 0.93 \\
Recall & \textbf{0.23} &  0.00 & \textbf{0.21} & \multicolumn{1}{c|}{0.00} & 0.42 & 0.00 & 0.42 & 0.00 & \textbf{0.04} & 0.52 & \textbf{0.12} & \multicolumn{1}{c|}{0.07} & -0.12 & 0.10 & -0.11 & 0.15 \\
F1 & 0.14 & 0.03 & 0.16 & \multicolumn{1}{c|}{0.02} &  \textbf{0.43} &  0.00  & \textbf{0.44} & 0.00 & -0.12 & 0.09 & -0.13 & \multicolumn{1}{c|}{0.10} & -0.08 & 0.32 & -0.11 & 0.10 \\ \hline\textbf{\underline{ROUGE-L}} & & & & \multicolumn{1}{c|}{} & & & & & & & & \multicolumn{1}{c|}{} &  & & & \\
Precision & -0.03 & 0.71 & 0.03 & \multicolumn{1}{c|}{0.66} & 0.20 & 0.00 & 0.31 & 0.00 & -0.15 & 0.06 & -0.18 & \multicolumn{1}{c|}{0.02} & \textbf{0.12} & 0.36 & \textbf{0.05} &  0.70 \\
Recall & \textbf{0.22} & 0.00 & \textbf{0.17} & \multicolumn{1}{c|}{0.02} & 0.36 & 0.00 & 0.36 & 0.00 & \textbf{0.09} & 0.22 & \textbf{0.13} & \multicolumn{1}{c|}{0.12} & -0.12 & 0.09 & -0.18 & 0.02 \\
F1 & 0.12 & 0.07 & 0.17 & \multicolumn{1}{c|}{0.06} &  \textbf{0.38} & 0.00 & \textbf{0.39} & 0.00 & -0.16 & 0.09 & -0.15 & \multicolumn{1}{c|}{0.07} & -0.08 & 0.21 & -0.11 & 0.11 \\ \hline\textbf{\underline{BERTScore}} & & & & \multicolumn{1}{c|}{} & & & & & & & & \multicolumn{1}{c|}{} &  & & & \\
Precision & 0.14 & 0.04 & 0.15 & \multicolumn{1}{c|}{0.03} & 0.36 & 0.00 & 0.43 & 0.00 & -0.13 & 0.00 & -0.13 & \multicolumn{1}{c|}{0.00} & \textbf{0.11} & 0.12 & \textbf{0.05} & 0.52 \\
Recall & 0.16 & 0.03 & 0.15 & \multicolumn{1}{c|}{0.04} & 0.34 & 0.00 & 0.33 & 0.00 & \textbf{0.02} & 0.78 & \textbf{0.03} & \multicolumn{1}{c|}{0.63} & -0.10 & 0.13 & -0.13 & 0.08 \\
F1 & \textbf{0.22} & 0.00 & \textbf{0.18} & \multicolumn{1}{c|}{0.00} & 0.38 & 0.00 & 0.38 & 0.00 & -0.09 & 0.12 & -0.10 & \multicolumn{1}{c|}{0.10} & 0.00 & 0.95 & -0.04 & 0.57 \\ \hline \thickhline
\end{tabular}%
\caption{Statistical Relationship between human judgement (relevance and factual consistency) and metric scores based on Pearson correlation and Spearman rank correlation coefficients and their p-values.}
\label{tab. rouge_bert_scores}}
\vspace{-0.3cm}
\end{table*}

\end{document}